%% file: anonymous-submission-latex-2027.tex
\documentclass[letterpaper]{article} 
\usepackage{aaai2027} 
\usepackage[hyphens]{url} 
\usepackage{graphicx} 
\usepackage{natbib} 
\usepackage{caption} 
\usepackage{algorithm}
\usepackage{algorithmic}
\usepackage{booktabs}
\usepackage{amsmath}
\usepackage{amssymb}

\title{MIND: Marginal-Invariant Neural Dependency Diffusion for Mixed-Type Tabular Generation}
\author{
Pengfei Li,
Mohammad Khalil
}

\affiliations{
Centre for the Science of Learning \& Technology (SLATE),\\
University of Bergen, Bergen, Norway
}

\begin{document}

\maketitle

\begin{abstract}
This paper proposes MIND, a marginal-invariant neural dependency diffusion model for mixed-type tabular data. MIND does not directly learn the joint distribution in the original heterogeneous feature space. Instead, it first maps different variable types into a unified latent dependency space via column-wise marginal transport. A conditional diffusion model then learns cross-column relationships. Copula-tangent denoising separates known marginal components from learnable dependency residuals. Rank projection during the sampling phase further mitigates marginal shift in reverse diffusion. Experiments across nine diverse tabular benchmarks show that MIND consistently improves marginal fidelity and dependency preservation over existing unified approaches. By explicitly isolating marginal modelling from dependency learning, MIND achieves a strong and stable balance among marginal fidelity, joint dependency preservation, and downstream prediction utility. This work supports separating marginal and dependency modelling as a principled and highly effective paradigm for complex mixed-type tabular generation.
\end{abstract}

\section{Introduction}
\input{sections/intro}

\section{Related Work}
\input{sections/background}

\section{Methodology}
\input{sections/methodology}

\section{Experiment Setup}

\input{sections/experiment_setup}

\section{Results}
\input{sections/results}

\section{Discussion}

\input{sections/discussion}

\section{Conclusion}
\label{sec:conclusion}
This paper proposes MIND, a marginal-invariant neural dependency diffusion model for mixed-type tabular data. MIND maps heterogeneous variables into a unified latent space through column-wise marginal transport, allowing conditional diffusion to focus on cross-column dependencies. Copula-tangent denoising and rank projection further reduce marginal drift during generation. 

Across diverse datasets, MIND achieves a strong balance among marginal fidelity, dependency preservation, and downstream utility. These results support explicit marginal-dependency decoupling as a practical design principle for mixed-type tabular generation.

\bibliography{references}


\end{document}

%% file: sections/intro.tex
Driven by the surging demand for data augmentation and privacy protection, synthetic tabular generation has become a critical research direction in machine learning \cite{borisov2022deep}. The current evaluation paradigm has evolved from singular macroscopic similarity to multidimensional metrics like downstream utility, statistical fidelity, and privacy risks \cite{stoian2025survey}. Unlike the uniform metric space of images or the natural sequential dependencies of text, tabular data is a complex combination of multivariate heterogeneous random variables \cite{grinsztajn2022tree,borisov2022deep}. Individual variables exhibit entirely distinct statistical behaviours (e.g., skewed continuous values, high-cardinality discrete values, or non-random missingness), and they intertwine with strong domain-specific nonlinear dependencies and explicit constraints \cite{xu2019modeling,zhao2021ctab}. Therefore, the key challenge in high-quality tabular generation lies in simultaneously achieving the statistical fidelity of single-column marginal distributions and the semantic validity of multi-column joint distributions.

Existing deep tabular generators do address feature heterogeneity, but mainly as a representation or optimisation problem rather than by separating marginal modelling from dependency learning. CTGAN and TVAE use conditional sampling, mode-specific normalisation, and reconstruction objectives to accommodate mixed-type columns \cite{xu2019modeling}. TabDDPM applies type-specific diffusion processes to numerical and categorical variables \cite{kotelnikov2023tabddpm}, while STaSy improves score-based generation through dedicated training strategies \cite{kim2022stasy}. TabSyn moves diffusion into a continuous latent space learned by a VAE \cite{zhang2024mixed}, and TabDiff introduces a joint mixed-type diffusion process with feature-wise noise schedules \cite{shi2025tabdiff}. These designs substantially improve tabular synthesis, yet column marginals and cross-column dependencies are still learned through shared representations and coupled objectives. 

This unified modelling paradigm requires models to balance column-level marginal distributions and cross-column joint structures under an end-to-end objective. Existing research alleviates this difficulty from various angles. For instance, CTGAN designs mode-specific normalisation and conditional sampling for multimodal continuous columns and imbalanced discrete categories. TabDDPM extends diffusion models to heterogeneous tabular data composed of continuous and discrete features. TabDiff further emphasises the challenges of complex inter-column correlations and fine-grained column-level distributions in mixed-type tabular generation \cite{xu2019modeling,kotelnikov2023tabddpm,shi2025tabdiff}. However, the fidelity of column-level marginal distributions cannot substitute for structural consistency at the pairwise, conditional, or full-joint levels \cite{yang2024structured}. When models fail to capture highly skewed continuous columns, long-tail categories, or missing patterns, the generated data often weakens the coverage of low-frequency subgroups and introduces biases in marginal distributions or inter-column dependencies \cite{grinsztajn2022tree,xu2019modeling,yang2024structured,shi2025tabdiff}. Therefore, we argue that the difficulty in tabular generation is not solely insufficient generator capacity. It also involves the representation-and-optimisation coupling dilemma when aligning marginal distributions and joint structures within a unified representation space.

To address these challenges, we propose a novel modelling perspective. Mixed-type tabular generation should explicitly decouple column marginal distributions and inter-column dependencies rather than fitting them simultaneously in the original heterogeneous feature space or an entangled continuous latent space. Based on this idea, we introduce \textbf{MIND}, the \textbf{M}arginal-\textbf{I}nvariant \textbf{N}eural \textbf{D}ependency diffusion model. MIND first maps continuous variables, categorical variables, and missing patterns into a unified normalised dependency space via column-wise marginal transformations. A neural diffusion model then learns the complex nonlinear and high-order dependencies. Inverse transformations finally generate valid mixed-type tabular data. Isolating heterogeneous marginal processing allows MIND to focus its model capacity entirely on cross-column dependency modelling.

\begin{figure*}[t]
    \centering
    \includegraphics[width=0.92\textwidth]{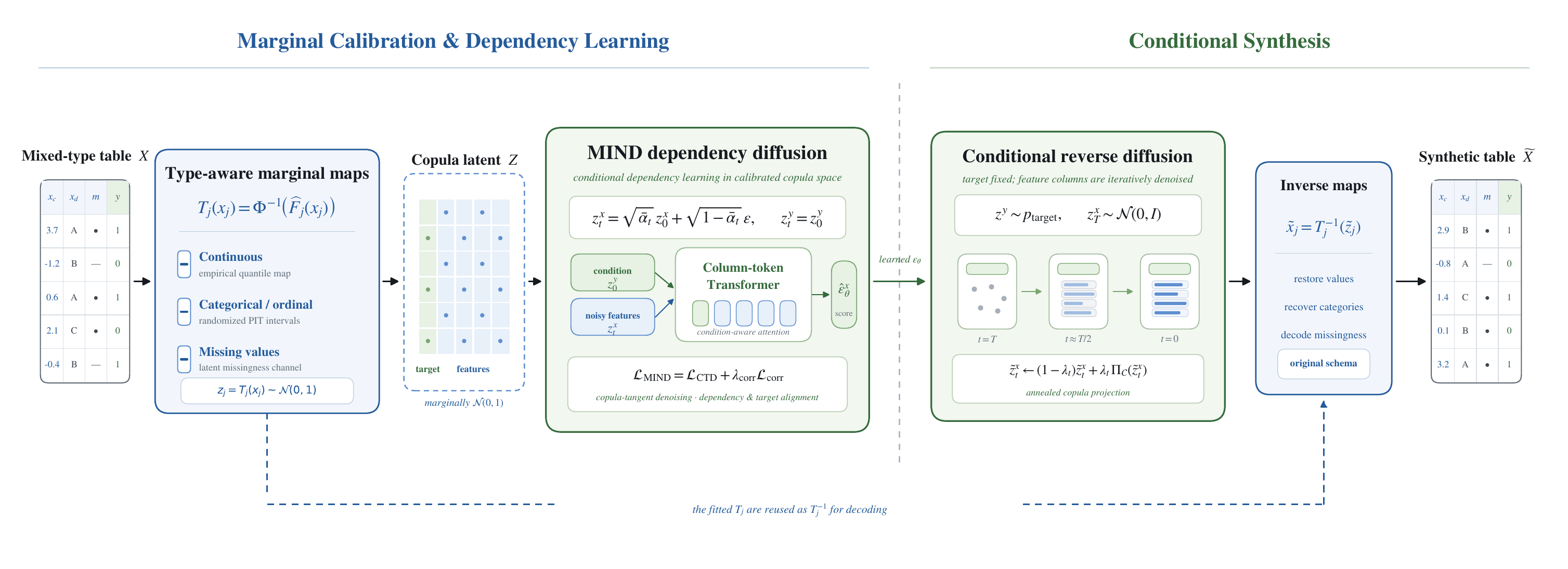}
    \caption{Overview of MIND: type-aware marginal mapping, conditional dependency diffusion, and calibrated inverse generation.}
    \label{fig:mind_methodology}
\end{figure*}

MIND is inspired by classic copula theory. Sklar's theorem decomposes any multivariate joint distribution into univariate marginal distributions and a copula that describes variable dependencies \cite{sklar1959fonctions,nelsen2006introduction}. Traditional copula models rely on predefined parametric families or complex structural selections. They struggle to capture nonlinear, high-order, and mixed-type dependencies in high-dimensional tabular data. MIND combines this classic decomposition with neural generative models. Column-wise transformations model heterogeneous marginals, and a diffusion model learns flexible dependency structures within a unified copula-normalised space. Crucially, rather than treating this mapping as a mere preprocessing step, MIND deeply customises the generative process to preserve this decoupled geometry rigorously. MIND is therefore not a simple modification of existing models. It is a marginal-invariant generation framework centred on dependency modelling. Figure~\ref{fig:mind_methodology} illustrates the MIND architecture.

The main contributions of this paper are as follows:

\begin{itemize}
\item We propose a marginal-invariant dependency modelling paradigm for mixed-type tabular generation, where column marginals are handled separately from cross-column dependence. MIND maps heterogeneous columns into a unified latent space through column-wise marginal transforms and learns the remaining dependency structure with a neural generative model. This decoupling allows the generation backbone to focus on statistical relationships rather than repeatedly fitting heterogeneous marginal shapes.

\item We introduce a dependency diffusion model incorporating copula-aware denoising in this latent space, latent alignment, and progressive marginal projection. These mechanisms capture high-order relationships while suppressing marginal shift to align dependency learning with marginal calibration.

\item We systematically compare MIND against classic statistical and deep tabular generative models on multiple public mixed-type datasets. We validate the framework's effectiveness through marginal fidelity, dependency structure fidelity, and downstream task utility.
\end{itemize}

%% file: sections/background.tex
\subsection{Synthetic Tabular Data Generation}

Synthetic tabular data generation methods broadly cover traditional probabilistic models, deep generative models, and sequence-based modelling approaches \cite{stoian2025survey}. Probabilistic models like Bayesian networks and copulas characterise variable relationships via explicit structural assumptions, offering some interpretability. However, their reliance on strong distributional or dependency assumptions often hinders them from adequately capturing complex joint distributions in high-dimensional, nonlinear, and mixed-type scenarios \cite{zhang2017privbayes,patki2016synthetic,xu2019modeling}. Subsequently, GAN and VAE-based methods are widely used for tabular synthesis. To address multimodal continuous columns and imbalanced discrete columns, CTGAN proposes designs like mode-specific normalisation and training-by-sampling \cite{xu2019modeling}. Other works introduce structural or sequential inductive biases. For instance, GOGGLE explicitly learns variable relationship graphs and models inter-column dependencies via message passing \cite{liu2023goggle}. GReaT and REaLTabFormer serialise tabular rows into text and generate records using autoregressive language models or GPT-style Transformers \cite{borisov2022language,solatorio2023realtabformer}, while TabMT models tabular feature generation with a masked Transformer \cite{gulati2023tabmt}. These methods enhance tabular generation from statistical structure, neural generation, and sequence modelling perspectives. Yet, most directly model the joint distribution in the original feature or encoding space, leaving marginal distribution fitting and inter-column dependency learning to the same generative process.

\subsection{Diffusion and Score-Based Generative Models for Tabular Data}

Recent tabular generators increasingly adopt diffusion or score-based models. TabDDPM directly models heterogeneous numerical and categorical features \cite{kotelnikov2023tabddpm}, while STaSy improves score-based tabular generation through dedicated stabilisation strategies \cite{kim2022stasy}. TabSyn performs diffusion in a VAE latent space \cite{zhang2024mixed}, and TabDiff introduces feature-wise diffusion processes for mixed-type variables \cite{shi2025tabdiff}. These methods improve generation quality but still learn marginal variation and cross-column dependence within a shared feature or latent representation. MIND instead performs diffusion in a copula-normalised space, separating marginal calibration from dependency learning.

\subsection{Copula Dependency Modelling}

Copula theory provides a classical statistical foundation for separating marginal distributions and dependency structures. By Sklar's theorem, a multivariate joint distribution can be decomposed into univariate marginal distributions and a copula function \cite{sklar1959fonctions,nelsen2006introduction}. This perspective highly aligns with the variable-level structure of tabular data, where each column usually possesses independent semantics and distribution shapes. Gaussian and vine copulas are representative methods in synthetic tabular data \cite{tagasovska2019copulas,sun2019learning}. Such methods naturally distinguish column-level marginal distributions from cross-column dependencies but usually rely on predefined copula families or structure selection. Thus, they require strong modelling assumptions when facing complex nonlinear dependencies, high-dimensional relations, and mixed-type data \cite{sun2019learning,fan2017high,genest2007primer}. We inherit this marginal-dependency decomposition perspective but adopt more flexible neural generative modelling within a normalised dependence space.

%% file: sections/methodology.tex
\subsection{Overview}
\label{sec:overview}

Let $\mathcal{D}=\{(\mathbf{x}_i,y_i)\}_{i=1}^{N}$ be a mixed-type tabular dataset. The vector $\mathbf{x}_i$ contains continuous, categorical, ordinal, and potentially missing feature variables, while $y_i$ is a designated fully observed target column. MIND addresses target-aware synthesis by first encoding each record through column-wise marginal transports:
\begin{equation}
(\mathbf{z}_0,z_0^y)
=
\mathcal{T}(\mathbf{x},y,\boldsymbol{\xi}),
\end{equation}
where $\boldsymbol{\xi}$ denotes dequantization randomness for discrete variables. The latent joint distribution is then factorised as
\begin{equation}
p(\mathbf{z}_0,z_0^y)
=
\widehat q_y(z_0^y)\,
p_{\theta}(\mathbf{z}_0\mid z_0^y),
\label{eq:factorization}
\end{equation}
where $\widehat q_y$ is the empirical target marginal prior after transport. The transports calibrate column marginals. The conditional diffusion model learns the feature dependency distribution given the sampled target.

\subsection{Marginal-Invariant Representation}
\label{sec:transport}

For variable $j$, MIND constructs
\begin{equation}
z_j
=
T_j(x_j,\xi_j)
=
\Phi^{-1}\!\left(U_j(x_j,\xi_j)\right),
\label{eq:transport}
\end{equation}
where $\Phi$ is the standard normal cumulative distribution function and $U_j\in(0,1)$ is a type-specific empirical quantile coordinate. Continuous variables use empirical mid-ranks and monotone interpolation. Categorical, binary, and ordinal variables are assigned disjoint quantile intervals according to smoothed empirical frequencies and are dequantized within the corresponding interval. This maps heterogeneous columns to approximately Gaussian-calibrated marginals, while inverse transports return valid values in the original domain \cite{sklar1959fonctions,nelsen2006introduction,dunn1996randomized}.

For each variable with missing values, MIND additionally introduces a binary missingness coordinate. Numerical values at missing positions are replaced by auxiliary Gaussian values and excluded from the denoising loss. Categorical missing values are represented by a dedicated token. Full transport and missingness constructions are provided in the appendix.

\subsection{Copula-Tangent Dependency Diffusion}
\label{sec:ctd}

MIND diffuses only non-target coordinates and keeps the target coordinate clean:
\begin{equation}
\mathbf{z}_t
=
a_t\mathbf{z}_0+b_t\boldsymbol{\epsilon},
\qquad
z_t^y=z_0^y,
\qquad
\boldsymbol{\epsilon}\sim
\mathcal{N}(\mathbf{0},\mathbf{I}),
\label{eq:forward}
\end{equation}
where $t\in\{1,\ldots,T\}$, $a_t=\sqrt{\bar\alpha_t}$, and $b_t=\sqrt{1-\bar\alpha_t}$ \cite{ho2020denoising,nichol2021improved}. In the normalised latent space, the noise admits the decomposition
\begin{equation}
\boldsymbol{\epsilon}
=
b_t\mathbf{z}_t+a_t\mathbf{r}_t,
\qquad
\mathbf{r}_t
=
\frac{\boldsymbol{\epsilon}-b_t\mathbf{z}_t}{a_t}.
\label{eq:residual_decomp}
\end{equation}
Under ideal standard-normal marginals, $r_{t,j}$ is uncorrelated with $z_{t,j}$. Thus, $b_t\mathbf{z}_t$ is an analytically determined marginal component, and the predictable part of $\mathbf{r}_t$ carries target-conditional dependency information.

Let $\mathcal{P}_{\boldsymbol{\eta}}(\cdot,\mathbf{z}_t,\mathbf{M})$ denote the type-aware copula-tangent operator defined in the appendix. It removes masked batch means and centred latent scale directions, with a relaxed projection for dequantized discrete coordinates. Given
\[
\mathbf{u}_{\theta}
=
f_{\theta}(\mathbf{z}_t,t,z_0^y),
\]
MIND uses the hybrid parameterisation
\begin{equation}
\widehat{\boldsymbol{\epsilon}}_{\theta}
=
\begin{cases}
\mathbf{u}_{\theta},
& t\leq\tau_{\mathrm{low}},\\[1mm]
b_t\mathbf{z}_t
+
a_t\mathcal{P}_{\boldsymbol{\eta}}
(\mathbf{u}_{\theta},\mathbf{z}_t,\mathbf{M}),
& t>\tau_{\mathrm{low}}.
\end{cases}
\label{eq:hybrid_ctd}
\end{equation}
The low-noise branch directly predicts total noise to preserve local details and discrete interval boundaries. The remaining steps predict a tangent-constrained dependency residual.

Define
\begin{equation}
\mathbf{D}_t
=
\begin{cases}
\boldsymbol{\epsilon}-\mathbf{u}_{\theta},
& t\leq\tau_{\mathrm{low}},\\[1mm]
\mathcal{P}_{\boldsymbol{\eta}}
(\mathbf{r}_t,\mathbf{z}_t,\mathbf{M})
-
\mathcal{P}_{\boldsymbol{\eta}}
(\mathbf{u}_{\theta},\mathbf{z}_t,\mathbf{M}),
& t>\tau_{\mathrm{low}}.
\end{cases}
\end{equation}
The denoising objective is
\begin{equation}
\mathcal{L}_{\mathrm{CTD}}
=
\mathbb{E}\!\left[
\frac{
\|\mathbf{M}\odot\mathbf{D}_t\|_F^2
}{
\|\mathbf{M}\|_1
}
\right],
\label{eq:ctd_loss}
\end{equation}
where $\mathbf{M}\in\{0,1\}^{B\times d}$ masks numerical coordinates that were originally missing.

\subsection{Noise-Adaptive Dependency Network}
\label{sec:network}

MIND represents each latent coordinate as a column token:
\begin{equation}
\mathbf{h}^{(0)}_j
=
E_{\mathrm{val}}(z_{t,j})
+
\mathbf{e}^{\mathrm{col}}_j
+
\mathbf{e}^{\mathrm{type}}_j
+
\mathbf{e}^{\mathrm{role}}_j
+
\mathbf{e}^{\mathrm{time}}_t.
\end{equation}
A column-wise Transformer models interactions among these tokens \cite{vaswani2017attention}.

To stabilise attention under heavy corruption, MIND interpolates between topology-based and value-based queries and keys:
\begin{equation}
\begin{aligned}
\mathbf{Q}_t
&=
\omega_t\mathbf{Q}^{\mathrm{top}}
+
(1-\omega_t)\mathbf{Q}^{\mathrm{val}},\\
\mathbf{K}_t
&=
\omega_t\mathbf{K}^{\mathrm{top}}
+
(1-\omega_t)\mathbf{K}^{\mathrm{val}}.
\end{aligned}
\label{eq:qk_mix}
\end{equation}
where
\begin{equation}
\omega_t
=
\frac{
\sqrt{1-\bar\alpha_t}
}{
\sqrt{1-\bar\alpha_T}
}.
\end{equation}
High-noise attention relies primarily on stable column identities. Low-noise attention increasingly uses sample-specific values. A time-dependent target-attention bias, an asymmetric condition mask, target-anchor reinjection, and a bounded column-tied readout further preserve the conditioning signal. Architectural details are given in the appendix.

\subsection{Objective and Copula-Projected Sampling}
\label{sec:objective_sampling}

MIND reconstructs the clean latent variables as
\begin{equation}
\widehat{\mathbf{z}}_0
=
\frac{
\mathbf{z}_t-b_t\widehat{\boldsymbol{\epsilon}}_{\theta}
}{
a_t
}.
\end{equation}
It then matches normal-score correlations among generated coordinates and between generated coordinates and the target. These terms act as a marginal-invariant second-order dependency anchor. Nonlinear and higher-order structure is learned by the diffusion Transformer \cite{liu2009nonparanormal}. Denoting the two terms by $\mathcal{L}_{\mathrm{GG}}$ and $\mathcal{L}_{\mathrm{GY}}$, the total objective is
\begin{equation}
\mathcal{L}_{\mathrm{MIND}}
=
\mathcal{L}_{\mathrm{CTD}}
+
\lambda_{\mathrm{dep}}
\left(
\mathcal{L}_{\mathrm{GG}}
+
\lambda_y\mathcal{L}_{\mathrm{GY}}
\right).
\label{eq:total_loss}
\end{equation}
Their exact forms are provided in the appendix.

For generation, MIND samples $\widetilde z^y\sim\widehat q_y$, initializes $\widetilde{\mathbf{z}}_T\sim\mathcal{N}(\mathbf{0},\mathbf{I})$, and performs target-conditional DDPM sampling. To correct accumulated marginal drift, it intermittently applies the column-wise rank projection
\begin{equation}
\left[\Pi_B(\mathbf{Z})\right]_{ij}
=
\Phi^{-1}\!\left(
\frac{r_{ij}}{B+1}
\right),
\label{eq:rank_projection}
\end{equation}
where $r_{ij}$ is the within-column rank of $Z_{ij}$. This monotone projection preserves empirical ranks while recalibrating each latent marginal. The projection schedule is detailed in the appendix.

\begin{algorithm}[t]
\caption{Training and Sampling with MIND}
\label{alg:mind}
\begin{algorithmic}[1]
\REQUIRE Training data $\mathcal{D}$ and diffusion horizon $T$
\ENSURE Synthetic table $\widetilde{\mathcal{D}}$
\STATE Fit transports $\mathcal{T}$ and encode $(\mathbf{Z}_0,\mathbf{z}_0^y)$
\WHILE{not converged}
    \STATE Sample $(\mathbf{z}_0,z_0^y)$, $t\in\{1,\ldots,T\}$, and $\boldsymbol{\epsilon}$
    \STATE Form $\mathbf{z}_t=a_t\mathbf{z}_0+b_t\boldsymbol{\epsilon}$ and keep $z_t^y=z_0^y$
    \STATE Compute $\mathbf{u}_{\theta}=f_{\theta}(\mathbf{z}_t,t,z_0^y)$
    \STATE Form $\widehat{\boldsymbol{\epsilon}}_{\theta}$ using Eq.~\eqref{eq:hybrid_ctd}
    \STATE Update $\theta$ by minimizing Eq.~\eqref{eq:total_loss}
\ENDWHILE
\STATE Sample $\widetilde z^y\sim\widehat q_y$ and initialize $\widetilde{\mathbf{z}}_T\sim\mathcal{N}(\mathbf{0},\mathbf{I})$
\FOR{$t=T,\ldots,1$}
    \STATE Perform one target-conditional reverse-diffusion update
    \STATE Apply the scheduled hard or soft rank projection
\ENDFOR
\STATE Decode $\widetilde{\mathcal{D}}=\mathcal{T}^{-1}(\widetilde{\mathbf{z}}_0,\widetilde z^y)$
\STATE \textbf{return} $\widetilde{\mathcal{D}}$
\end{algorithmic}
\end{algorithm}

%% file: sections/experiment_setup.tex
We evaluate MIND on six public classification benchmarks:
Adult~\cite{becker1996adult}, Default Credit
Card~\cite{yeh2009default}, FICO HELOC~\cite{fico2018heloc},
Covertype~\cite{blackard1998covertype}, Online
Shoppers~\cite{sakar2018shoppers}, and Telco
Churn~\cite{ibmtelco}. We further include two public regression
benchmarks: Beijing PM2.5~\cite{chen2015beijing}, which predicts
hourly PM2.5 concentration from temporal and meteorological variables,
and Online News Popularity~\cite{fernandes2015online}, which predicts
the number of social-media shares from article-level features.
These datasets extend the evaluation to continuous targets from
environmental and media domains. We also use a controlled HeavyTail
stress test containing heavy-tailed numerical variables, a long-tail
categorical attribute, and structured missingness. Covertype is
evaluated on a class-stratified 50,000-row subset for computational
tractability~\cite{blackard1998covertype}.

To rigorously assess generation quality, we select baselines that represent a comprehensive spectrum of direct statistical and deep generative modelling paradigms. While recent autoregressive models sequence tables into text, we focus on continuous latent and feature-space generators: an independent empirical-marginal sampler (Indep.) and Gaussian Copula~\cite{patki2016synthetic} as classical approximations, CTGAN and TVAE~\cite{xu2019modeling} as standard deep tabular generators, and TabSyn~\cite{zhang2024mixed} and TabDiff~\cite{shi2025tabdiff} as the recent state-of-the-art diffusion-based methods. For each seed in $\{42,43,44\}$, all methods use the same stratified $64\%/16\%/20\%$ train/validation/test split and generate $|\mathcal{D}_{\mathrm{train}}|$ synthetic records.

We assess marginal fidelity using the Kolmogorov--Smirnov statistic~\cite{massey1951kolmogorov}, Wasserstein distance~\cite{villani2009optimal}, total variation, and Jensen--Shannon divergence~\cite{lin1991divergence}; dependency preservation using Pearson/Spearman correlation-matrix and mutual-information errors~\cite{cover2006elements}; and predictive utility using train-synthetic-test-real (TSTR) AUC, accuracy, and macro-F1~\cite{esteban2017realvalued}. Distributional fidelity and coverage are further evaluated by $\alpha$-Precision and $\beta$-Recall~\cite{alaa2022faithful}, while real--synthetic distinguishability is measured by classifier two-sample-test (C2ST) AUC~\cite{lopezpaz2017revisiting} and propensity-score mean squared error (pMSE)~\cite{snoke2018general}. Results are reported as mean $\pm$ standard deviation over three seeds. Complete implementation and evaluation details are provided in the supplementary.

%% file: sections/results.tex
\newcommand{\best}[1]{\textbf{#1}}

\subsection{Overall Generation Quality}
\label{sec:overall_results}

\begin{table*}[t]
\centering

{%
\fontsize{9pt}{10pt}\selectfont
\setlength{\tabcolsep}{1pt}
\renewcommand{\arraystretch}{1.08}

\begin{tabular*}{\textwidth}{
  @{\extracolsep{\fill}}
  lcccccccc
  @{}
}
\toprule
Metric
& Indep.
& G-Copula
& CTGAN
& TVAE
& TabSyn
& TabDiff
& MIND
& \shortstack{Sig.\\vs.} \\
\midrule

TSTR score $\uparrow$
& 0.344$\pm$0.293
& 0.620$\pm$0.322
& 0.586$\pm$0.317
& 0.647$\pm$0.345
& 0.576$\pm$0.545
& 0.634$\pm$0.406
& \best{0.693$\pm$0.335}
& I \\

KS $\downarrow$
& 0.008$\pm$0.005
& 0.008$\pm$0.005
& 0.147$\pm$0.059
& 0.118$\pm$0.039
& 0.033$\pm$0.022
& 0.028$\pm$0.028
& \best{0.003$\pm$0.005}
& C,V,S,D \\

JS $\downarrow$
& 0.012$\pm$0.014
& 0.012$\pm$0.014
& 0.080$\pm$0.029
& 0.119$\pm$0.080
& 0.027$\pm$0.017
& 0.019$\pm$0.013
& \best{0.005$\pm$0.012}
& C,V,S,D \\

Column JS $\downarrow$
& 0.015$\pm$0.011
& 0.017$\pm$0.011
& 0.101$\pm$0.035
& 0.118$\pm$0.050
& 0.050$\pm$0.033
& 0.043$\pm$0.039
& \best{0.009$\pm$0.009}
& C,V,S,D \\

Pearson err. $\downarrow$
& 0.104$\pm$0.046
& 0.037$\pm$0.013
& 0.064$\pm$0.023
& 0.059$\pm$0.029
& 0.017$\pm$0.005
& \best{0.014$\pm$0.006}
& 0.016$\pm$0.005
& I,G,C,V \\

Pairwise MI err. $\downarrow$
& 0.067$\pm$0.058
& 0.045$\pm$0.040
& 0.033$\pm$0.018
& 0.033$\pm$0.022
& 0.010$\pm$0.007
& 0.009$\pm$0.009
& \best{0.008$\pm$0.005}
& I,G,C,V \\

C2ST gap $\downarrow$
& 0.450$\pm$0.112
& 0.412$\pm$0.149
& 0.462$\pm$0.054
& 0.466$\pm$0.029
& 0.238$\pm$0.176
& 0.214$\pm$0.192
& \best{0.167$\pm$0.120}
& I,G,C,V \\

pMSE $\downarrow$
& \best{0.003$\pm$0.004}
& 0.004$\pm$0.004
& 0.052$\pm$0.028
& 0.088$\pm$0.070
& 0.019$\pm$0.030
& 0.020$\pm$0.032
& 0.003$\pm$0.004
& C,V,S,D \\

$\alpha$-Precision $\uparrow$
& 0.727$\pm$0.262
& 0.862$\pm$0.167
& 0.881$\pm$0.163
& \best{0.988$\pm$0.014}
& 0.981$\pm$0.009
& 0.983$\pm$0.010
& 0.981$\pm$0.006
& I,G,C \\

$\beta$-Recall $\uparrow$
& 0.839$\pm$0.165
& 0.926$\pm$0.079
& 0.927$\pm$0.074
& 0.944$\pm$0.018
& 0.978$\pm$0.008
& \best{0.983$\pm$0.008}
& 0.981$\pm$0.007
& I,G,C,V \\

\bottomrule
\end{tabular*}
}

\caption{Aggregate generation quality across nine datasets.
Bold denotes the best mean in each row. The ``Sig. vs.'' column lists
baselines significantly outperformed by MIND under two-sided paired
Wilcoxon signed-rank tests on nine dataset-level seed means, with Holm
correction over six comparisons within each metric
($p_{\mathrm{adj}}<0.05$).
I denotes Indep., G denotes G-Copula, C denotes CTGAN, V denotes TVAE,
S denotes TabSyn, and D denotes TabDiff.}
\label{tab:main-quality}

\end{table*}

Table~\ref{tab:main-quality} summarises marginal fidelity, dependency preservation, distinguishability, and support coverage over nine datasets. MIND records the best mean on six of the ten metrics. Its KS, JS, and Column JS errors are 0.003, 0.005, and 0.009. Relative to the next lowest means, these errors fall by 62.5\%, 58.3\%, and 40.0\%.

MIND also obtains the lowest Pairwise MI error at 0.008 and the lowest C2ST gap at 0.167, compared with 0.214 for TabDiff. Its Pearson error is 0.016, behind TabDiff at 0.014 but slightly ahead of TabSyn at 0.017. Support coverage remains high, with both $\alpha$-Precision and $\beta$-Recall at 0.981. TVAE is best on precision, while TabDiff is best on recall. The independent model gives the lowest pMSE, yet its TSTR score and dependency errors are much worse. This contrast shows that pMSE should be read together with utility and dependency measures.

\subsection{Downstream Utility}
\label{sec:utility_results}

\begin{table*}[t]
\centering
{\small
\setlength{\tabcolsep}{1mm}
\renewcommand{\arraystretch}{1.06}
\begin{tabular}{@{}lccccccc@{}}
\toprule
Dataset & Indep. & G-Copula & CTGAN & TVAE & TabSyn & TabDiff & MIND \\
\midrule
Adult
& 0.500 $\pm$ 0.030
& 0.792 $\pm$ 0.012
& 0.886 $\pm$ 0.001
& 0.884 $\pm$ 0.005
& 0.905 $\pm$ 0.004
& \best{0.910 $\pm$ 0.004}
& 0.899 $\pm$ 0.003 \\

Beijing$^*$
& -0.001 $\pm$ 0.003
& 0.233 $\pm$ 0.031
& 0.166 $\pm$ 0.010
& 0.186 $\pm$ 0.108
& 0.529 $\pm$ 0.022
& \best{0.560 $\pm$ 0.033}
& 0.525 $\pm$ 0.028 \\

Covertype
& 0.502 $\pm$ 0.047
& 0.721 $\pm$ 0.013
& 0.625 $\pm$ 0.069
& 0.882 $\pm$ 0.010
& 0.565 $\pm$ 0.010
& 0.594 $\pm$ 0.004
& \best{0.940 $\pm$ 0.001} \\

Default
& 0.471 $\pm$ 0.008
& 0.693 $\pm$ 0.007
& 0.721 $\pm$ 0.016
& 0.734 $\pm$ 0.011
& 0.753 $\pm$ 0.010
& 0.750 $\pm$ 0.009
& \best{0.756 $\pm$ 0.008} \\

FICO
& 0.529 $\pm$ 0.016
& 0.779 $\pm$ 0.004
& 0.638 $\pm$ 0.056
& 0.785 $\pm$ 0.006
& 0.784 $\pm$ 0.007
& 0.785 $\pm$ 0.002
& \best{0.788 $\pm$ 0.004} \\

HeavyTail
& 0.452 $\pm$ 0.064
& 0.718 $\pm$ 0.015
& 0.607 $\pm$ 0.051
& 0.726 $\pm$ 0.015
& 0.727 $\pm$ 0.002
& \best{0.737 $\pm$ 0.005}
& 0.723 $\pm$ 0.021 \\

News$^*$
& -0.307 $\pm$ 0.464
& -0.076 $\pm$ 0.095
& \best{-0.038 $\pm$ 0.082}
& -0.070 $\pm$ 0.077
& -0.834 $\pm$ 0.730
& -0.396 $\pm$ 0.445
& -0.137 $\pm$ 0.214 \\

Shoppers
& 0.467 $\pm$ 0.127
& 0.881 $\pm$ 0.014
& 0.846 $\pm$ 0.007
& 0.874 $\pm$ 0.016
& 0.915 $\pm$ 0.006
& \best{0.921 $\pm$ 0.008}
& 0.900 $\pm$ 0.008 \\

Telco
& 0.483 $\pm$ 0.078
& 0.835 $\pm$ 0.011
& 0.819 $\pm$ 0.010
& 0.821 $\pm$ 0.026
& 0.841 $\pm$ 0.012
& \best{0.846 $\pm$ 0.016}
& 0.843 $\pm$ 0.019 \\
\midrule
Avg.
& 0.344 $\pm$ 0.293
& 0.620 $\pm$ 0.322
& 0.586 $\pm$ 0.317
& 0.647 $\pm$ 0.345
& 0.576 $\pm$ 0.545
& 0.634 $\pm$ 0.406
& \best{0.693 $\pm$ 0.335} \\
\bottomrule
\end{tabular}%
}
\caption{Per-dataset utility under the train-on-synthetic,
test-on-real protocol. Scores are averaged over XGBoost, LightGBM,
and MLP, using AUC for classification datasets and $R^2$ for
regression datasets (marked with $^*$). Entries report mean $\pm$ standard deviation
over three seeds. The Avg. row averages dataset-level means.
Higher is better, and bold denotes the best mean in each row.}
\label{tab:main-utility-tstr}
\end{table*}

Table~\ref{tab:main-utility-tstr} reports TSTR utility on seven classification datasets and two regression datasets. MIND achieves the highest average score of 0.693, exceeding TVAE, TabDiff, and TabSyn by 0.046, 0.059, and 0.117. On Beijing, TabDiff leads with 0.560, while MIND reaches 0.525 and remains close to TabSyn at 0.529. News is more difficult. Every method has a negative mean $R^2$, with CTGAN best at $-0.038$ and MIND at $-0.137$.

On the classification datasets, MIND ranks first on Covertype, Default, and FICO, and second on Telco. Its largest gain appears on Covertype, where it reaches 0.940 compared with 0.882 for TVAE. On Adult, HeavyTail, and Shoppers, the gaps to the best method are 0.011, 0.014, and 0.021. The gap on Telco is 0.003. Additionally, MIND shows better cross-seed stability than Tabdiff and Tabsyn.

\subsection{Dependency Preservation}
\label{sec:dependency_results}

\begin{figure}[t]
\centering
\includegraphics[width=\columnwidth]{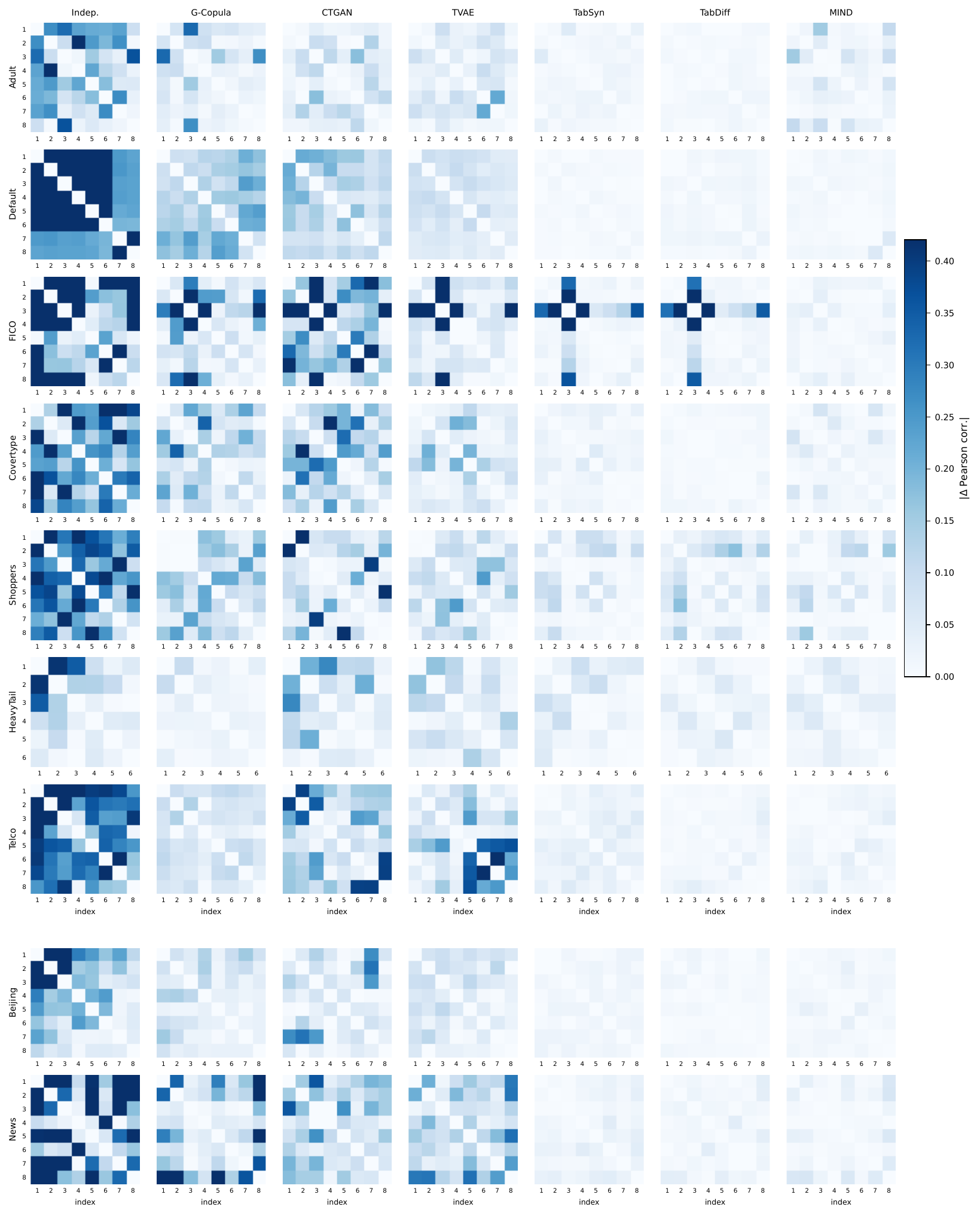}
\caption{Pairwise Pearson correlation errors across datasets. Lighter colours indicate more accurate dependency preservation.}
\label{fig:pairwise-corr-error}
\end{figure}

Figure~\ref{fig:pairwise-corr-error} further illustrates error distributions across variable pairs. The independent marginal model forms large high-error areas across most datasets. This shows that accurate univariate recovery cannot reconstruct joint structures. Gaussian Copula improves significantly but still leaves concentrated error blocks in FICO, Covertype, and Shoppers. This reflects the limits of fixed dependency families on complex mixed-type relationships. CTGAN and TVAE errors show strong dataset dependence with noticeable deviations in certain pairs.

TabSyn, TabDiff, and MIND exhibit lighter overall error distributions. MIND specifically reduces locally concentrated high errors in Default, FICO, and Telco. It achieves balanced dependency recovery across different pairs. Meanwhile, TabDiff retains a slightly lower overall Pearson error. This aligns with the aggregated results in Table~\ref{tab:main-quality}. Thus, the heatmap illustrates that MIND avoids severe distortion in specific local dependencies rather than strictly outperforming TabDiff on all pairs.

\subsection{Marginal Distribution Analysis}
\label{sec:marginal_results}

\begin{figure}[t]
\centering
\includegraphics[width=\columnwidth]{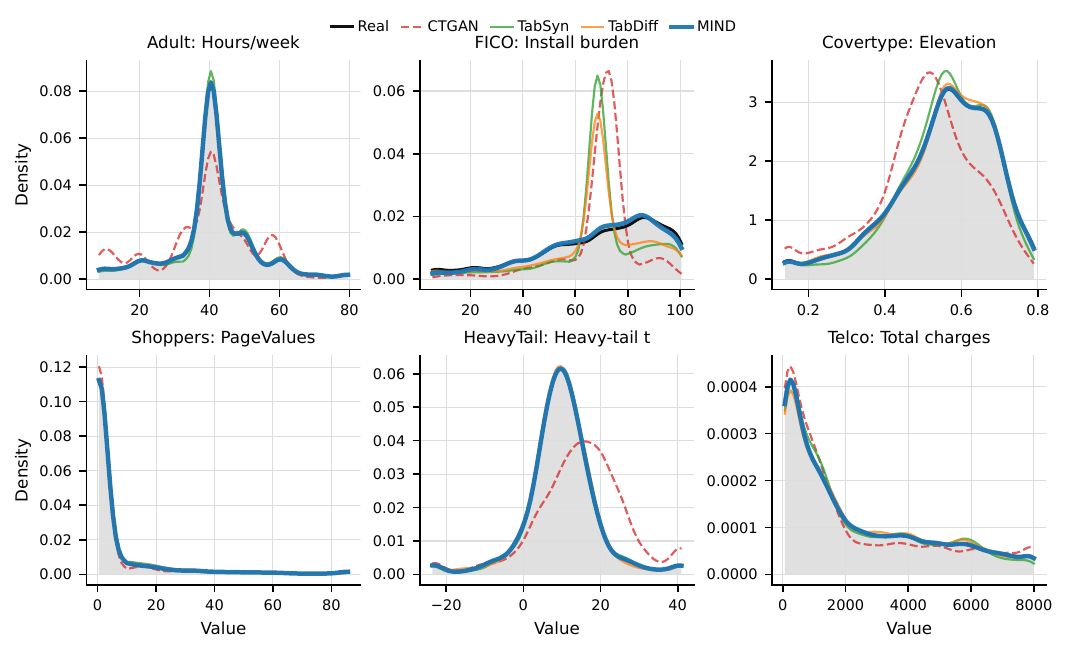}
\caption{Single-column density overlays for representative continuous variables (seed 42). The real distribution is shown in black. Selected neural baselines are compared with MIND, and each x-axis is clipped to the range between the real-data 1st and 99th percentiles for readability.}
\label{fig:column-density}
\end{figure}

Figure~\ref{fig:column-density} compares the density estimates of six representative continuous variables. MIND successfully recovers the sharp main peak of Hours/week in Adult, the broad peak and right shoulder of Install burden in FICO, and the asymmetric peak of Elevation in Covertype. In contrast, CTGAN exhibits varying degrees of peak shift on these variables. TabSyn and TabDiff fit well overall but still show deviations on
certain narrow peaks or multi-scale structures. TabSyn, TabDiff, and CTGAN all hallucinate a sharp spurious mode around 65–75 that doesn't exist in the real density, while MIND tracks the true flat/bimodal shape.

This difference becomes more pronounced on challenging distributions. For HeavyTail, MIND accurately recovers the centre, peak width, and right-tail decay of the true distribution. Conversely, CTGAN produces a significantly right-shifted and over-dispersed density. For the highly skewed Shoppers PageValues and the long-tailed Telco Total charges, MIND preserves the high-density region near zero and the tail decay as values increase.

\subsection{Ablation Study}
\label{sec:ablation}

\begin{table*}[t]
\centering
{\small
\setlength{\tabcolsep}{4pt}
\renewcommand{\arraystretch}{1.06}
\begin{tabular}{@{}lccccc@{}}
\toprule
Variant
& TSTR $\uparrow$
& \shortstack{Column\\JS $\downarrow$}
& \shortstack{Pairwise MI\\error $\downarrow$}
& \shortstack{C2ST\\gap $\downarrow$}
& $\beta$-Recall $\uparrow$ \\
\midrule

Full MIND
& 0.844 $\pm$ 0.085
& \textbf{0.0135 $\pm$ 0.0109}
& \textbf{0.0082 $\pm$ 0.0041}
& \textbf{0.140 $\pm$ 0.104}
& 0.9801 $\pm$ 0.0024 \\

w/o CTD
& 0.843 $\pm$ 0.087
& 0.0136 $\pm$ 0.0109
& 0.0090 $\pm$ 0.0038
& 0.144 $\pm$ 0.103
& 0.9791 $\pm$ 0.0023 \\

w/o Corr.
& 0.841 $\pm$ 0.089
& 0.0136 $\pm$ 0.0109
& 0.0089 $\pm$ 0.0038
& 0.144 $\pm$ 0.104
& 0.9797 $\pm$ 0.0023 \\

w/o Attn. Extras
& \textbf{0.845 $\pm$ 0.082}
& 0.0136 $\pm$ 0.0109
& 0.0089 $\pm$ 0.0038
& 0.144 $\pm$ 0.100
& 0.9795 $\pm$ 0.0032 \\

w/o Projection
& 0.844 $\pm$ 0.089
& 0.0432 $\pm$ 0.0084
& 0.0091 $\pm$ 0.0031
& 0.146 $\pm$ 0.115
& \textbf{0.9813 $\pm$ 0.0020} \\

\bottomrule
\end{tabular}%
}
\caption{Component ablation of MIND over three datasets and three
seeds. Entries report mean $\pm$ standard deviation over nine runs.
Higher is better for TSTR and $\beta$-Recall, while lower is better
for the remaining metrics. Bold denotes the best mean in each column.}
\label{tab:ablation}
\end{table*}

Table~\ref{tab:ablation} shows that copula projection is the key component for maintaining marginal fidelity. Removing this module increases Column JS from 0.0135 to 0.0432, an approximate 3.2-fold increase. Removing CTD, dependency regularisation, or attention enhancements has a minor impact on coverage, but each removal raises the C2ST gap from 0.140 to about 0.144. This indicates that these components jointly improve the realism of the overall joint distribution. Notably, removing the correlation regularisation increases the Pairwise MI error from 0.0082 to 0.0089 and slightly reduces the TSTR score. This confirms that explicitly regularising normal-score correlations not only stabilises dependency learning but also yields better multivariate mutual information and downstream utility. Overall, the full MIND achieves the best performance in marginal fidelity, pairwise dependency preservation, and distinguishability, while maintaining highly competitive downstream utility and coverage.

%% file: sections/discussion.tex
Compared with the strongest baselines in our experiments, MIND does not dominate every metric or dataset. MIND leads on Covertype, Default, and FICO, and obtains lower average marginal, mutual-information, and C2ST errors. Although the aggregate TSTR mean favours MIND, this difference is influenced substantially by Covertype. We therefore interpret the results as showing that MIND is competitive with diffusion-based SOTA. Separating marginals from dependence also has a clear precedent in Gaussian and vine copula synthesis \cite{patki2016synthetic,sun2019learning}. MIND differs from these methods by replacing a fixed copula family with target-conditional neural diffusion in a normalised dependence space, while correcting marginal drift during sampling.

Covertype illustrates the practical effect of this design. TabSyn and TabDiff obtain TSTR scores of 0.565 and 0.594, compared with 0.940 for MIND. Although TabDiff achieves a lower pairwise Pearson error, the Covertype stress-case analysis shows that it has a substantially larger C2ST gap and a pronounced shift in target-class mass (Supplementary Table 6). Class conditioning is also considered by CTGAN and TabDDPM, while TabSyn models the target jointly with other columns in a learned latent space \cite{xu2019modeling,kotelnikov2023tabddpm,zhang2024mixed}. MIND instead samples the target from its empirical marginal prior and learns $p_\theta(\mathbf{x}\mid y)$. The diffusion model therefore does not need to reconstruct class proportions, which helps account for its lower target-marginal error on Covertype.

The independent marginal model obtains low pMSE and small marginal errors, yet performs poorly in TSTR and dependency preservation. On Covertype, TabDiff achieves a lower Pearson error than MIND but substantially worse TSTR and C2ST results. Matching linear normal-score dependence therefore does not by itself ensure downstream utility or overall distributional similarity. MIND consequently uses normal-score correlation only as an auxiliary regularizer, while the diffusion model learns broader dependency structures. This evaluation across complementary fidelity and utility criteria follows recent systematic frameworks for synthetic tabular data assessment \cite{du2025systematic,yang2024structured}.

The ablation results provide the clearest evidence for the sampling projection. Removing it increases Column JS from 0.0135 to 0.0432, corresponding to an approximately threefold degradation in marginal fidelity. Removing CTD, correlation regularisation, or the attention additions changes the C2ST gap from 0.140 to between 0.144 and 0.146. These differences are small relative to the reported variation and do not establish interaction effects among the components. Removing correlation regularisation also changes Pairwise MI error from 0.0082 to 0.0089 and TSTR from 0.844 to 0.841. The ablation therefore strongly supports the role of projection in marginal calibration, while the aggregate evidence for the remaining components is more modest.

MIND comes with limitations. Currently, we assume a designated target column and perform batch-level generation, which limits direct use in target-free or multi-target settings. Its rank-based calibration also depends on sufficiently large generation batches. Future work will extend the framework to more flexible conditioning schemes, batch-independent sampling, and privacy-aware training, while preserving the separation between marginal modelling and dependency learning.